\documentclass[letterpaper, 10 pt, conference]{ieeeconf}  

\IEEEoverridecommandlockouts
\usepackage{cite}
\usepackage[export]{adjustbox}
\usepackage{amsmath,amssymb,amsfonts}
\usepackage{mathtools}
\usepackage{subfigure}
\usepackage{graphicx}
\usepackage{soul}
\usepackage{textcomp}
\usepackage{xcolor}
\usepackage{verbatim}
\usepackage{float}
\usepackage{algorithmicx}
\usepackage{algorithm}
\usepackage{algpseudocode}
\usepackage[caption=false]{subfig}
\def\BibTeX{{\rm B\kern-.05em{\sc i\kern-.025em b}\kern-.08em
    T\kern-.1667em\lower.7ex\hbox{E}\kern-.125emX}}

\graphicspath{{figures/}}
\sethlcolor{cyan}

\title{\bf \LARGE Road Boundary Estimation Using Sparse Automotive Radar Inputs}

\author{
Aaron Kingery and Dezhen~Song
\thanks{A. Kingery and D. Song are with CSE Department, Texas A\&M University, College Station, TX 77843, USA. D. Song is also with the Robotics Department of Mohamed Bin Zayed University of Artificial Intelligence (MBZUAI) in Abu Dhabi, UAE. Email: \texttt{dezhen.song@mbzuai.ac.ae.}}
\thanks{This work was supported in part by General Motors/SAE Autodrive Challenge II.}
}

\begin{document}

\maketitle

\begin{abstract}
This paper presents a new approach to detecting road boundaries based on sparse radar signals. We model the roadway using a homogeneous model and derive its conditional predictive model under known radar motion. Using the conditional predictive model and model radar points using a Dirichlet Process Mixture Model (DPMM), we employ Mean Field Variational Inference (MFVI) to derive an unconditional road boundary model distribution. 
In order to generate initial candidate solutions for the MFVI, we develop a custom Random Sample and Consensus (RANSAC) variant to propose unseen model instances as candidate road boundaries. For each radar point cloud we alternate the MFVI and RANSAC proposal steps until convergence to generate the best estimate of all candidate models. We select the candidate model with the minimum lateral distance to the radar on each side as the estimates of the left and right boundaries.
We have implemented the proposed algorithm in C++. We have tested the algorithm and it has shown satisfactory results. More specifically, the mean lane boundary estimation error is not more than 11.0 cm.
\end{abstract}

\section{Introduction}
Detection of road boundaries is important for Advanced Driver Assistance Systems (ADAS) and autonomous driving. Significant efforts have been dedicated toward their robust and accurate detection, including proposals to integrate radar reflectors into roadway infrastructure to make road boundaries more visible to the sensor \cite{clarke2016synthetic} \cite{feng2018lane}.  Vision and lidar approaches have been the popular sensor choices, including camera-radar fusion \cite{serfling2008road} \cite{patel2022road} and lidar-radar fusion \cite{homm2010efficient}, of which \cite{romero2021road} provides a survey. However, vision and lidar are easily affected by severe weather conditions, leading to poor performance. On the other hand, radars are less sensitive to environmental conditions. Developing a radar-based approach complements existing vision and lidar-based approaches and will offer a backup solution for vehicles. 

\begin{figure}[t!]
    \centering
    \includegraphics[width=0.4\linewidth, viewport=5 5 725 725, clip=true]{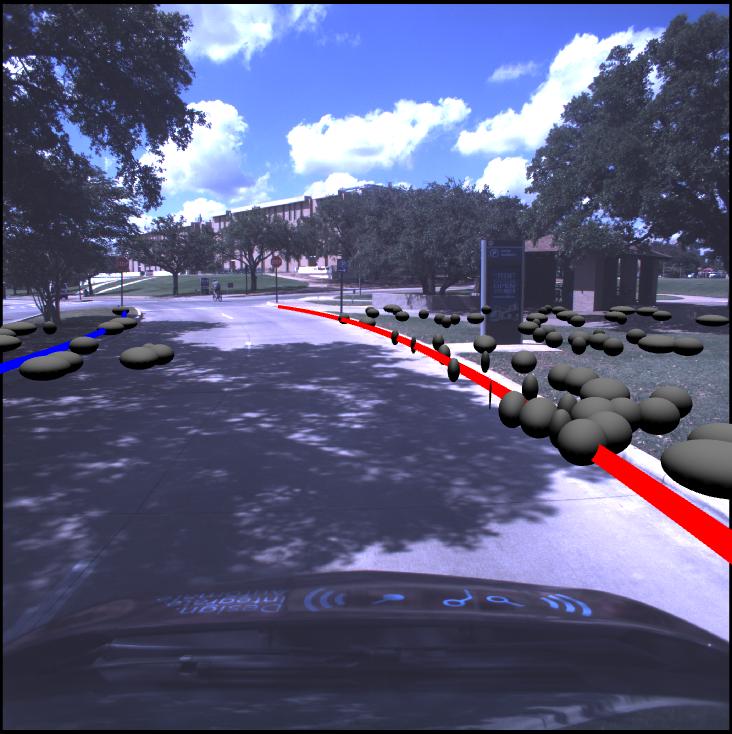}
    \hspace{0.2cm}
    \includegraphics[width=0.4\linewidth, viewport=0 0 913 912, clip=true]{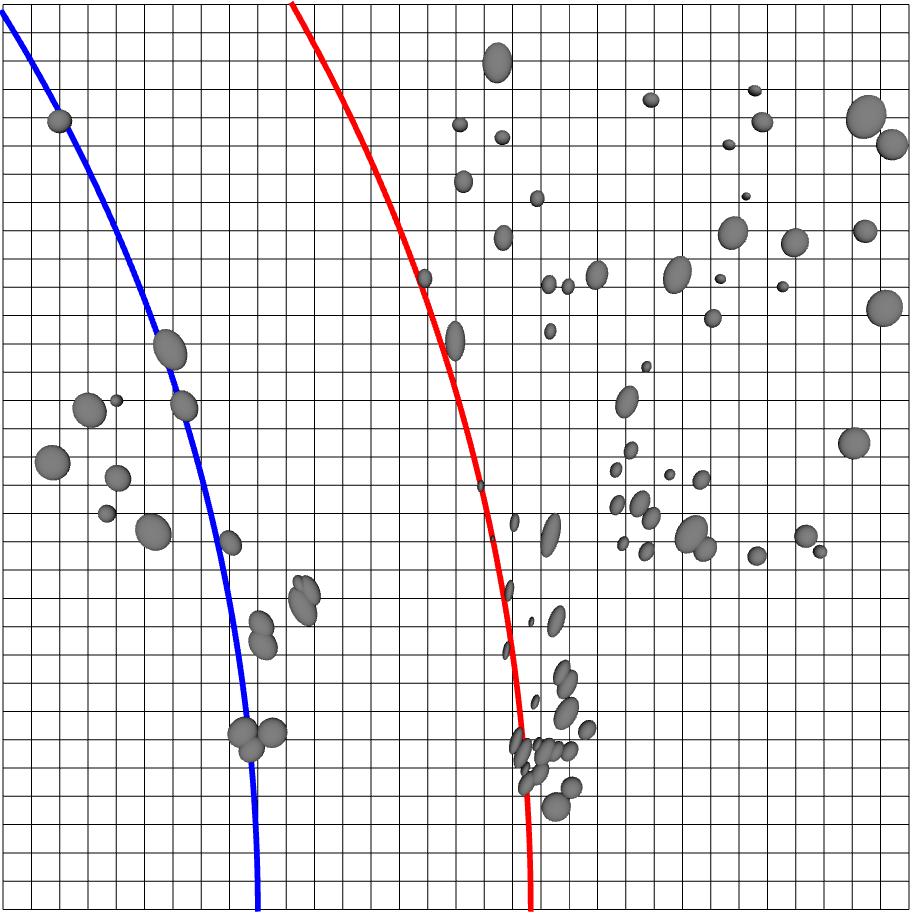}
    \caption{An example of the radar road boundary estimation problem. On the left, the scenario is shown from the camera perspective, with results from the radar frame projected onto the camera image. On the right, a top down orthographic perspective of the scene. Grey ellipses are radar target detections, and the blue and red curves are the left and right road boundaries, estimated from the radar, respectively. The grid has 1 meter spacing.}
    \label{fig:title_fig}
\end{figure}

The signals of a typical automotive radar are pretty sparse after target detection and hence are typically relegated only to the detection of dynamic objects such as other vehicles, bicycles, and pedestrians. In this paper, we propose a new method to detect road boundaries based on sparse radar signals. By sparse signals, we refer to the filtered output of common automotive radars instead of the raw reflectivity images that are only available to radar developers. Fig.~\ref{fig:title_fig} illustrates the radar inputs and our problem. 

We model the roadway using a homogeneous four-element parameterized arc model and derive its conditional predictive form under known radar motion. Using the conditional predictive model and model radar points using a Dirichlet Process Mixture Model (DPMM), we employ Mean Field Variational Inference (MFVI) to derive an unconditional road boundary model distribution. 
In order to generate initial candidate solutions for the MFVI, we develop a custom Random Sample and Consensus (RANSAC) variant to propose unseen model instances as candidate road boundaries. For each radar point cloud we alternate the MFVI and RANSAC proposal steps until convergence to generate the best estimate of all candidate models. We select the candidate model with the minimum lateral distance to the radar on each side as the estimates of the left and right boundaries.

We have implemented the proposed algorithm in C++ as a module in the Robot Operating System (ROS). We tested the algorithm on data collected from a Continental ARS430 automotive Doppler radar and it has shown satisfactory results. More specifically, the mean lane boundary estimation error is not more than 11.0 cm, which is reasonable when considering radar wavelength.

\section{Related Work}


In general, when a radar is used in robot/vehicle perception, there are, broadly, two forms of output to consider: radar reflectivity images and radar target detection point clouds.

Radar reflectivity images are analogous to a camera image but with cells instead of pixels and with cell dimensions typically including at least range and azimuth. 
For road boundary estimation in this case, Nikolova and Hero \cite{nikolova2000segmentation} model the boundaries in the polar coordinate space of the radar reflectivity image and identify them as the edges of continuous, homogeneous regions in the image with constant width and curvature. 
Kaliyaperumal et al. \cite{kaliyaperumal2001algorithm} propose a deformable model for the boundaries, a likelihood function to match the model with the edges in the imate, and use the Metropolis-Hastings algorithm with simulated annealing to find the optimal match.
Guo et al. \cite{guo2014road} present the stripe Hough transform which is capable of detecting lines in the reflectivity image when there are orthogonal deviations from the boundary. Werber et al. \cite{werber2019association} estimate straight line landmarks from the reflectivity image and associate them over time for vehicle localization. 

For most commercially available automotive radars, like the one used for this work, the reflectivity image is not an available output, instead the image is downsampled into a point cloud of target detections generally representing points of peak reflectivity in the image. 
This radar point cloud data, while similar in format to lidar data, is comparatively sparse, noisy, and unstructured. As a consequence, lidar-based boundary detection approaches are generally not applicable to radar point clouds. 
A common technique to overcome these issues for radar point clouds is to temporally `stack' the point clouds by transforming the points into a static coordinate frame using some known localization (e.g. GPS), merging them, and then estimating using this more dense stacked point cloud. 
Lundquist et al. \cite{lundquist2009estimation} present two such approaches, estimation from an occupancy grid map and estimation from a Quadratic Program (QP) with prior sorting of target detections into left and right sets and outlier rejection. 
Xu et al. \cite{xu2020road} generate an occupancy grid from temporally stacked radar observations and combine edge detection detection and RANSAC in order to identify linear boundaries. 

A significant weakness of the stacking based approach is that it relies upon robust and accurate localization over the stacking window. When the localization is not accurate, especially when traversing turns and corners, the stacked radar point cloud often exhibits a distinct `smearing' of the points which can degrade or destroy the estimation quality. 
For this reason, the algorithm we propose tracks the road boundaries over time, applying new observations to update the existing estimate in a Bayesian manner as opposed to the stacking based approach. 
Among such approaches, Lee et al. \cite{lee2018application} provide an instantaneous estimate of the road curvature and Lundquist et al. \cite{lundquist2009estimation} track points and quadratic curve segments along the roadway as extended objects via an Extended Kalman Filter (EKF), but neither approach estimates the road boundaries. 
Lundquist et al. \cite{lundquist2011road}, as a prior step towards radar road intensity mapping, perform a K-means regression clustering of cubic curves in highway scenarios. Compared with this work, our algorithm directly estimates the primary road boundaries themselves, allows a dynamic number of candidate tracks, and makes a probabilistic, as opposed to binary, assignment of target detections to tracks, giving the algorithm robustness in the presence of the obfuscating clutter common in non-highway scenarios.

\section{Problem Definition}
Consider that we have a vehicle equipped with automotive radar traveling along a roadway and we would like to estimate the location of the left and right road boundaries relative to the sensor at the time $t$ of each measurement. Let us define $\{\mathcal{D}_t\}$ as the dynamic radar ego coordinate system at time $t$ that moves with the radar so that the $X$-axis is forward and the $Y$-axis is to the right.

\subsection{Assumptions}
\begin{itemize}
    \item[a.1] The radar is mounted such that its forward direction is parallel with the vehicle's longitudinal axis.
    \item[a.2] The radar is positioned such that it is located between the left and right road boundaries.
    \item[a.3] The primary left and right road boundaries within the field of view (i.e. ignoring discontinuities at intersections, driveways, etc.) can be approximated as a circular arc or line in the Cartesian coordinate space.
    \item[a.4] The motion of the radar between two subsequent time steps is known or estimated. For this purpose, we use the instantaneous radar ego-velocity estimation \cite{kingery2022improving}, however, other odometry modalities (e.g. inertial) should also be sufficient. 
\end{itemize}

\subsection{Sensor Model}
\label{subsec:radar_inputs}
We will follow the sensor model in our previous work with these radars \cite{kingery2022improving}.  For completeness, we reiterate it here. Our sensor is a 77GHz automotive Doppler radar which periodically transmits a pulse and reports a set of the received reflections as estimated target detections. 
We use $\mathbf{Z}_t=\{\mathbf{z}_{t,1},\ldots,\mathbf{z}_{t,n}\}$ to denote the set of target detections received as a datagram from a transmitted radar pulse. 
We consider each detection $\mathbf{z}_{t,i}$ as a noisy observation of some ground truth measurement source $\mathbf{s}_{t,i}.$
Each target detection is reported in polar coordinates in the radar coordinate frame $\{\mathcal{D}_t\}$ and consists of the measurements.
\begin{equation}
    \mathbf{z}_{t,i}=
    \begin{bmatrix}
    r_{t,i} & \theta_{t,i}
    \end{bmatrix}^\mathsf{T} = \mathbf{s}_{t,i} + \mathbf{v}_{t,i},
\end{equation}
where, $i \in \{1,...,N \}$ indicates the target index, $r_{t,i}\in[0,r_{\max}]$ is the $i$-th target range, $\theta_{t,i}\in[\theta_{\min}, \theta_{\max}]$ is the $i$-th target azimuth where $\theta_{t,i}=0$ indicates the forward direction, $\theta_{t,i}<0$ indicates a target to the left, and $\theta_{t,i}>0$ indicates a target to the right, and $\mathbf{v}_{t,i}$ is the observation noise such that $\mathbf{v}_{t,i}\sim\text{Normal}(\mathbf{0},\mathbf{\Sigma}_{t,i})$ where
\(
\mathbf{\Sigma}_{t,i} = 
\begin{bmatrix}
\sigma_{r,i}^2 & 0 \\
0 & \sigma_{\theta,i}^2
\end{bmatrix}. 
\)
The Cartesian parameterization of the target position in $\{\mathcal{D}_t\}$ is defined in the usual manner,
\(
    \begin{bmatrix}
        x_{t,i} & y_{t,i}
    \end{bmatrix}^\mathsf{T}
    =
    \begin{bmatrix}
        r_{t,i}\cos(\theta_{t,i}) & r_{t,i}\sin(\theta_{t,i})
    \end{bmatrix}^\mathsf{T}.
    \label{eq:polar_conversion}
\)

\subsection{Road Boundary Model}
In the Cartesian coordinate space, roadways are generally constructed as a series of circular arcs connected by linear segments, which may be considered as circular arcs with infinite radius. We desire to use a single model which can represent both, and, for this reason, we make use of the quadratic representation of a conic with constant curvature, 
\begin{equation}
   \beta_1(x^2+y^2)+\beta_2 x+\beta_3 y+\beta_4=0,
    \label{eq:cartesian_road_boundary_model}
\end{equation} 
where $\beta_1\neq0$ represents a circle and $\beta_1=0$ a line.

As the radar makes observations in polar coordinates, it is desirable to express the model in polar coordinates as well so we have,
\begin{equation}
\beta_1 r^2+\beta_2 r\cos(\theta)+\beta_3 r\sin(\theta)+\beta_4=0.
\label{eq:polar_road_boundary_model}
\end{equation}
From this we define the road boundary as the set of points 
\begin{equation}
    \{\mathbf{s}
    \mid\boldsymbol{\beta}^\mathsf{T}\boldsymbol{\phi}(\mathbf{s}) = 0, \mathbf{s}\in[0,r_{\max}]\times[\theta_{\min},\theta_{\max}]\}
    \label{eq:road_boundary_model}
\end{equation}
where 
\(
    \mathbf{s}=
    \begin{bmatrix}
        r & \theta
    \end{bmatrix}^\mathsf{T}
\) is a point expressed in polar coordinates,
\(
    \boldsymbol{\phi}(\mathbf{s})=
    \begin{bmatrix}
        r^2 & r\cos(\theta) & r\sin(\theta) & 1
    \end{bmatrix}^\mathsf{T}
\)
are the model basis functions, and 
\(
    \boldsymbol{\beta}=
    \begin{bmatrix}
        \beta_1 & \beta_2 & \beta_3 & \beta_4
    \end{bmatrix}^\mathsf{T}
\) 
are the model coefficients. We note that the model is homogeneous, i.e. $\boldsymbol{\beta}\equiv c\boldsymbol{\beta},$ for all $c\in\mathbb{R}.$

\subsection{Problem Definition}
Given a series of radar datagrams $\mathbf{Z}_0,\ldots,\mathbf{Z}_t,$ we will estimate the left and right road boundaries at time $t$, $\boldsymbol{\beta}_{t,l}$ and $\boldsymbol{\beta}_{t,r},$ as the coefficients of our road boundary model.

\section{Algorithm}
\begin{figure*}[t!]
    \centering
    \includegraphics[width=0.79\linewidth, viewport=16 331 948 533, clip=true]{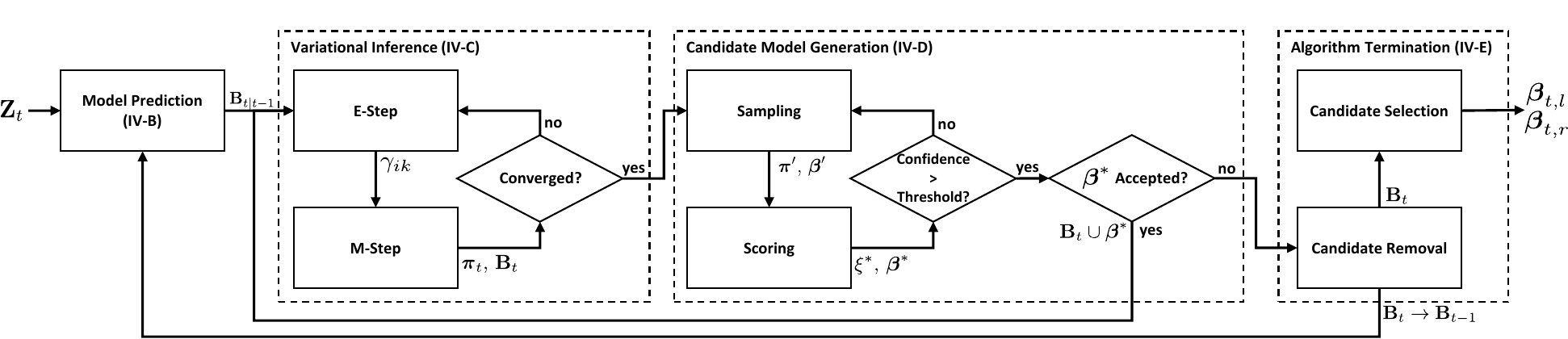}
    \caption{Algorithm Flow Chart.}
    \label{fig:algorithm}
\end{figure*}
Our algorithm operates in three primary phases: model prediction from known radar motion, model estimation alternating candidate updates from the radar observations and proposing new candidates, and algorithm termination where we output the boundary estimates and prepare for subsequent radar observations. We provide a graphical overview of the flow of our algorithm in Fig. \ref{fig:algorithm}.

\subsection{The Roadway as a Dirichlet Process Mixture Model} \label{subsec:the_roadway_as_a_dirichlet_process_mixture_model}
As described in \ref{subsec:radar_inputs}, we model each target detection $\mathbf{z}_{t,i}$ as the detection of some true source point $\mathbf{s}_{t,i}$ with zero-mean Gaussian observation noise $\mathbf{v}_{t,i}.$ We model the radar points as being generated by a DPMM, where the points are classified into two groups: outliers whose source is not necessarily modeled and inliers which are generated by a candidate model. Therefore, we will have $K+1$ models where model index $k=0$ indicates the outlier model and $k\in\{1,\ldots,K\}$ indicates a candidate model.  Each observed target $\mathbf{z}_{t,i}$ is associated with the latent variable $c_{t,i}\sim\text{Categorical}(\boldsymbol{\pi}_t)$ representing which model produced the target detection, where $\boldsymbol{\pi}_t=
\begin{bmatrix}
    \pi_{t,0} & \cdots & \pi_{t,K}\
\end{bmatrix}^\mathsf{T}\sim\text{Dirichlet}(\boldsymbol{\alpha}_t)$ are the mixture weights and $\boldsymbol{\alpha}_t=
\begin{bmatrix}
    \alpha_{t,0} & \cdots & \alpha_{t,K}
\end{bmatrix}^\mathsf{T}$ is the prior mixture concentration.

Ideally, one would represent the source point as its own latent variable described by a Spatial Distribution Model (SDM) on the source object. This is difficult in practice as such an SDM is heavily dependent upon the geometry of the scene and would need to account for occlusions, object shape, etc. Additionally, estimating the source point as a latent variable significantly complicates posterior inference unless the SDM is in the exponential family. Instead, we rely upon a Greedy Association Model (GAM) as in \cite{faion2014reducing}, and we will show that this produces a very straightforward algorithm for inference. The result is that we convert $\mathbf{z}_{t,i}$ into a pseudo-observation of the model $\boldsymbol{\beta}_{t,k}$ via the implicit shape function of our model $h(\mathbf{z}_{t,i},\boldsymbol{\beta}_{t,k})=\boldsymbol{\beta}_{t,k}^\mathsf{T}\boldsymbol{\phi}(\mathbf{z}_{t,i}),$ as in \cite{zhang1997parameter}. Propagating the measurement uncertainty, we can say that $h(\mathbf{z}_{t,i},\boldsymbol{\beta}_{t,k})\sim\text{Normal}(0,\sigma_{ik}^2)$ where $\sigma_{ik}^2=\boldsymbol{\beta}_{t,k}^\mathsf{T}\boldsymbol{\Phi}(\mathbf{z}_{t,i})\boldsymbol{\Sigma}_{t,i}\boldsymbol{\Phi}(\mathbf{z}_{t,i})^\mathsf{T}\boldsymbol{\beta}_{t,k}$ with $\boldsymbol{\Phi}(\mathbf{z}_{t,i})$ being the Jacobian of $\boldsymbol{\phi}(\mathbf{z}_{t,i}).$ We note that this GAM results in a biased estimation for curved shapes for which \cite{faion2014reducing} identifies and prescribes a correction. However, in the case of road boundary estimation with radar, the measurement uncertainty is not typically large enough relative to the curvature for this bias to have a significant effect. Thus, we consider $E[h(\mathbf{z}_{t,i},\boldsymbol{\beta}_{t,k})] = 0$ to be a reasonable approximation in this case. Finally, we model the outlier targets simply as observations from a uniform SDM over the field of view, $\mathbf{z}_{t,i}\mid c_{t,i}=0\sim\text{Uniform}([0,r_{\max}]\times[\theta_{\min},\theta_{\max}])$.

Due to the fact that our road boundary model is homogeneous, it is necessary to choose a scale for the coefficients. Setting the scale such that $\boldsymbol{\beta}_{t,k}^\mathsf{T}\boldsymbol{\beta}_{t,k}=1$ is a natural choice. Consequently, the coefficients of each candidate model is distributed such that $\boldsymbol{\beta}_{t,k}\sim\text{Bingham}(\mathbf{C}_{t,k}^{-1})$, which is the analogue to the normal distribution but conditioned on the unit hypersphere \cite{bingham1974antipodally}. The Bingham distribution has found related use in Quaternion Bingham Filters e.g. \cite{gilitschenski2015unscented,li2019geometry,wang2019comparison}, which provide more detail on the properties of the Bingham distribution. Note that the Bingham distribution is typically parameterized by the eigen decomposition $\mathbf{C}_{t,k}^{-1}=\mathbf{V}\mathbf{\Lambda}\mathbf{V}^\mathsf{T},$ however, for our purposes it will be more natural to work with $\mathbf{C}_{t,k}^{-1}$ directly. While ideally it should be the case that $p(\boldsymbol{\beta}_{t,k})=p(c\boldsymbol{\beta}_k)$ where $c\in\mathbb{R}$, for our distribution, it is only true that $p(\boldsymbol{\beta}_{t,k})=p(-\boldsymbol{\beta}_{t,k}),$ i.e. the distribution is antipodally symmetric, but it will serve as a reasonable approximation in return for useful mathematical properties, namely that the Bingham distribution is in the exponential family, making the inference more tractable. Finally, it is the case that $E[\boldsymbol{\beta}_{t,k}]=\mathbf{0},$ however, this trivial solution will not be useful for inference. Therefore, we will instead use the mode of the distribution which is the eigenvector corresponding to the minimum eigenvalue of the eigenvalue problem $\mathbf{C}_{t,k}^{-1}\boldsymbol{\beta}_{t,k}=\lambda\boldsymbol{\beta}_{t,k}.$ Thus, with some abuse of notation, we will say that $E[\boldsymbol{\beta}_{t,k}]$ is equal to this mode.

In summary, we consider the generative model of radar detections to be
\begin{itemize}
    \item $N_t$: Number of detections. 
    \item $K_t$: Number of candidate models. 
    \item $\boldsymbol{\alpha}_t$: prior concentration. 
    \item $\boldsymbol{\pi}_t \sim \text{Dirichlet}(\boldsymbol{\alpha}_t)$.
    \item $c_{t,i} \sim \text{Categorical}(\boldsymbol{\pi}_t)$.
    \item $\mathbf{z}_{t,i} \sim \text{Uniform}([0,r_{\max}]\times[\theta_{\min},\theta_{\max}])\;\text{if}\;c_{t,i}=0.$
    \item $\boldsymbol{\beta}_{t,k}^\mathsf{T}\boldsymbol{\phi}(\mathbf{z}_{t,i}) \sim \text{Normal}(0,\sigma_{ik}^2)\;\text{if}\;c_{t,i}=\{1,\ldots,K_t\}.$
    \item $\boldsymbol{\beta}_{t,k} \sim \text{Bingham}(\mathbf{C}_{t,k}^{-1}).$    
\end{itemize}

\subsection{Model Prediction}
\label{subsec:model_prediction}
Let us say that from time $t-1$ to $t$ the radar frame is rotated about its vertical axis by $\Delta_{\psi}$ and its position is translated 
\(
\begin{bmatrix}
    \Delta_x & \Delta_y
\end{bmatrix}^\mathsf{T}
\)
then the transformation matrix for the radar coordinate frame is 
\begin{equation}
    \mathbf{T}_t =
    \begin{bmatrix}
        \cos\Delta_\psi & -\sin\Delta_\psi & \Delta_x \\
        \sin\Delta_\psi & \cos\Delta_\psi & \Delta_y \\
        0 & 0 & 1
    \end{bmatrix},
\end{equation}
and a static point is predicted to be transformed in the radar coordinate frame such that 
\begin{equation}
    \begin{bmatrix}
        x_{t\mid t-1} \\
        y_{t\mid t-1} \\
        1
    \end{bmatrix}
    =
    \mathbf{T}_t^{-1}
    \begin{bmatrix}
        x_{t-1} \\
        y_{t-1} \\
        1
    \end{bmatrix}.
\end{equation}
While the Bingham distribution is only closed under orthonormal transformations, the transition can instead be applied to the underlying normal distribution. Applying the transformation to given road boundary model coefficients $\boldsymbol{\beta}_{t-1}$ and applying additive noise $\mathbf{Q}_t$ results in
\begin{equation}
    \boldsymbol{\beta}_{t\mid t-1}=\mathbf{F}_t\boldsymbol{\beta}_{t-1} + \mathbf{Q}_t
\end{equation}
where
\begin{equation}
\begin{aligned}
    \mathbf{F}_t &= \\
    &\begin{bmatrix}
        1 & 0 & 0 & 0 \\
        2(\Delta_x\cos\Delta_\psi+\Delta_y\sin\Delta_\psi) & \cos\Delta_\psi & \sin\Delta_\psi & 0 \\
        2(\Delta_y\cos\Delta_\psi-\Delta_x\sin\Delta_\psi) & -\sin\Delta_\psi & \cos\Delta_\psi & 0 \\
        \Delta_x^2 + \Delta_y^2 & \Delta_x & \Delta_y & 1
    \end{bmatrix}.
\end{aligned}
\end{equation}
Thus the predicted road boundary model becomes
\begin{equation}
    \boldsymbol{\beta}_{t\mid t - 1}\sim\text{Bingham}((\mathbf{F}_t\mathbf{C}_{t-1,k}\mathbf{F}_{t}^\mathsf{T} + \mathbf{Q}_t)^{-1}).
\end{equation}

\subsection{Variational Inference}
\label{subsec:variational_inference}
While Markov Chain Monte Carlo (MCMC) methods, especially Gibbs sampling, are often used to estimate the posterior of such models, in an online application, the uncertainty of the burn-in period and mixing quality makes their convergence unreliable. Instead, we will use Mean Field Variational Inference 
(MFVI) in order to estimate the posterior distribution of the DPMM, at a small cost to accuracy in return for much more reliable convergence \cite{blei2006variational}.

For simplicity of notation, let us drop subscript $t$ from the variables $\mathbf{c}_t$, $\boldsymbol{\pi}_t$, and $\mathbf{B}_t$. The objective of MFVI then is to find an approximating joint distribution $q$ of the true joint distribution $p$ such that
\begin{equation}
    q(\mathbf{c},\boldsymbol{\pi},\mathbf{B})\approx p(\mathbf{Z},\mathbf{c},\boldsymbol{\pi},\mathbf{B}),
\end{equation}
where $q(\mathbf{c},\boldsymbol{\pi},\mathbf{B})=q(\mathbf{c})q(\boldsymbol{\pi})q(\mathbf{B})$ and $p(\mathbf{Z},\mathbf{c},\boldsymbol{\pi},\mathbf{B})=p(\mathbf{Z}\mid\mathbf{c},\mathbf{B})p(\mathbf{c}\mid\boldsymbol{\pi})p(\boldsymbol{\pi})p(\mathbf{B}).$ In practice, this is solved by minimizing the KL-Divergence between $p$ and $q$. For a given factor, the log of the optimal approximating distribution $q^*$ is known to be proportional to the expected value of the log of the true joint distribution $p$ over the other factors of the distribution. For example, $\log q^*(\mathbf{c})\propto E_{\boldsymbol{\pi},\mathbf{B}}[\log p(\mathbf{Z},\mathbf{c},\boldsymbol{\pi},\mathbf{B})].$ For this latent variable model, the result is an iterative procedure analogous to the Expectation-Maximization (EM) algorithm. In each iteration we will first compute an expectation step (E-Step), followed by a maximization step (M-Step). Iteration proceeds until the parameter estimation converges. A derivation of the E-Step and M-Step is provided in Appendices \ref{apx:estep_deriv} and \ref{apx:mstep_deriv} respectively.

\textbf{E-Step:} 
We solve for $q^*(\mathbf{c})$ which gives,
\begin{equation}
\label{eq:E_c}
    \gamma_{ik} = E_\mathbf{c}[\,[c_i=k]\,] = \frac{\rho_{ik}}{\sum_{j=0}^K \rho_{ij}}
\end{equation}
where 
$[c_i=k]$ is the Iverson bracket which is equal to one when $c_i=k$ and zero otherwise
and
\begin{equation}
    \rho_{ik} = E_{\boldsymbol{\pi}}[\pi_k] 
    \begin{cases}
        \mathcal{U}(\mathbf{z}_i;[0,r_{\max}]\times[\theta_{\min},\theta_{\max}]) & \text{if}\;k = 0 \\
        \mathcal{N}(\hat{\boldsymbol{\beta}}_k^\mathsf{T}\boldsymbol{\phi}(\mathbf{z}_i); 0, \sigma_{ik}^2) & \text{if}\;k>0
    \end{cases}
\end{equation}
with $\hat{\boldsymbol{\beta}}_k=E[\boldsymbol{\beta}_{k}]$ (see end of Sec. \ref{subsec:the_roadway_as_a_dirichlet_process_mixture_model}) and $\mathcal{U}$ and $\mathcal{N}$ representing uniform and normal probability density functions respectively. Equation \eqref{eq:E_c} is also equivalent to $p(c_i=k\mid \mathbf{z}_i).$

\textbf{M-Step:} 
We are then able to solve for the other parameters, $q^*(\boldsymbol{\pi})$ which gives
\begin{equation}
    E_{\boldsymbol{\pi}}[\pi_k]=\frac{\alpha_k + \sum_{i=1}^N\gamma_{ik}}{\sum_{j=0}^K \left(\alpha_j + \sum_{i=1}^N\gamma_{ij}\right)},
\end{equation}
and $q^*(\mathbf{B})$ which gives 
\begin{equation}
    \boldsymbol{\beta}_k\sim\text{Bingham}\left(\mathbf{C}_k^{-1} + \sum_{i=1}^N\gamma_{ik}\frac{\boldsymbol{\phi}(\mathbf{z}_i)\boldsymbol{\phi}(\mathbf{z}_i)^\mathsf{T}}{\sigma_{ik}^2}\right).
\end{equation}

\subsection{Candidate Model Generation}
\label{subsec:candidate_model_generation}
Due to the fact that the MFVI converges towards local optima and not necessarily the global optimum, the quality of the results is heavily dependent upon the initial conditions. Therefore, it is necessary to generate high quality initial estimates of the candidate models to be included in the inference. In order to accomplish this, we take inspiration from the structure of instantaneous multi-model fitting approaches, especially Progressive-X \cite{barath2019progressive}, wherein, a RANSAC model proposal step and a model optimization step are alternately applied. In our case, the model optimization is accomplished via the MFVI and the model proposal is accomplished by a custom RANSAC variant. A single iteration of our RANSAC variant algorithm consists of two steps: sampling and scoring.

\textbf{Sampling:} 
We instantiate a proposal model $\boldsymbol{\beta}'$ exactly from a sample of 3 detections i.e. 4 coefficients minus 1 for homogeneity. The random sample is generated from the Reservoir Sampling algorithm \cite{efraimidis2006weighted}, wherein each detection $\mathbf{z}_i$ is included in the sample with probability proportional to $\gamma_{i0}$ i.e. the probability that $\mathbf{z}_i$ belongs to the outlier class. Let $\mathbf{B}'=\mathbf{B}\cup\boldsymbol{\beta}'.$ Additionally, Let 
\(
\boldsymbol{\alpha}' =
\begin{bmatrix}
    \boldsymbol{\alpha}^\mathsf{T} & 3
\end{bmatrix}^\mathsf{T}
\), then $\boldsymbol{\pi}'=\frac{\boldsymbol{\alpha}'}{\|\boldsymbol{\alpha}'\|_1},$ where $\|\cdot\|_1$ is the L1 norm.

\textbf{Scoring:} 
Given $\mathbf{B}'$ and $\boldsymbol{\pi}'$ we compute a single E-Step as described by equation \eqref{eq:E_c} for $\gamma_{ik}'.$ We define the proposal score as the difference in the expected number of points in the outlier class, given the proposal,
\begin{equation}
    \xi = \sum_{i=1}^N \gamma'_{i0} - \gamma_{i0}.
\end{equation}
We maintain the best proposal $\boldsymbol{\beta}^*$ which has the greatest score $\xi^*.$

The confidence of the best proposal at iteration $j$ is
\begin{equation}
    1 - \left(1 - \left(\frac{\xi^*}{\sum_{i=1}^N \gamma_{i0}}\right)^3\right)^j,
\end{equation}
and we terminate the RANSAC algorithm when the confidence is greater than some threshold e.g. 99\%. If $\xi^*$ is greater than some acceptance threshold, then $\boldsymbol{\beta}^*$ is added to the set of candidate models. The acceptance threshold may be tuned depending upon how conservative we would like to be in adding new candidate models wherein the higher the threshold, the less likely we are to accept the proposal. Generally, this threshold should be greater than the minimum sample size of 3, otherwise it is nearly guaranteed to accept the proposal. 

\subsection{Algorithm Termination}
\label{subsec:algorithm_termination}
Given that the view of the scene changes with time, we cannot simply use the posterior model concentration $\boldsymbol{\alpha}_t$ as the prior for the next radar measurement at timestep $t+1$. Instead, we let the next prior be a moving average of the expected number of points assigned to each model,
\begin{equation}
    \alpha_{t,k} = (1-c)\alpha_{t-1,k}+ c\sum_{i=1}^N\gamma_{ik},
\end{equation}
where $c\in[0, 1]$ controls how strongly the current measurement affects the concentration prior wherein $c=1$ indicates that the current iteration completely determines the prior for the next iteration and $c=0$ indicates that we maintain a constant prior assigned in candidate generation.

Given $\alpha_{t,k},$ we eliminate any models which no longer sufficiently contribute to the DPMM. Simply, if $\alpha_{t,k}$ is below a chosen threshold, the candidate model $k$ is removed from the DPMM, otherwise, the values of $\alpha_{t,k},$ $\mathbf{C}_{t,k}^{-1},$ and $\boldsymbol{\beta}_{t,k}$ are used as a prior for the next radar measurement.

Lastly, we choose the candidates to be output as $\boldsymbol{\beta}_{t,l}$ and $\boldsymbol{\beta}_{t,r}.$ Given we are attempting to recognize the primary road boundaries to the left and right, we expect them to intercept the radar y-axis. We separate candidates into left and right groups depending on the sign of this y-intercept and select one from each group with y-intercept nearest to the radar origin to be $\boldsymbol{\beta}_{t,l}$ and $\boldsymbol{\beta}_{t,r}$ respectively.

\section{Experiments}
\begin{figure}[t!]
\centering
\subfigure[]{\includegraphics[width=0.29\linewidth, viewport=5 5 725 725, clip=true]{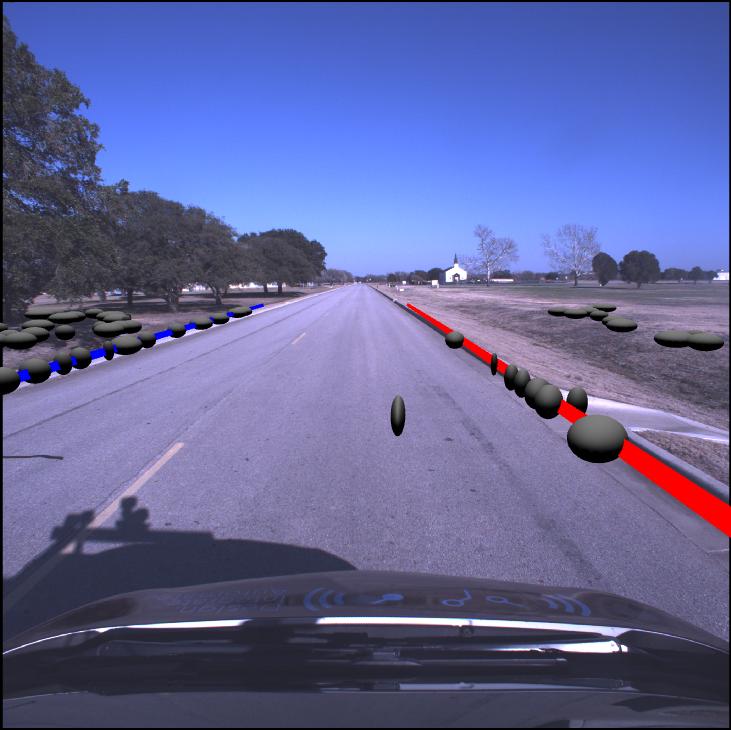}}
\subfigure[]{\includegraphics[width=0.29\linewidth, viewport=0 0 913 912, clip=true]{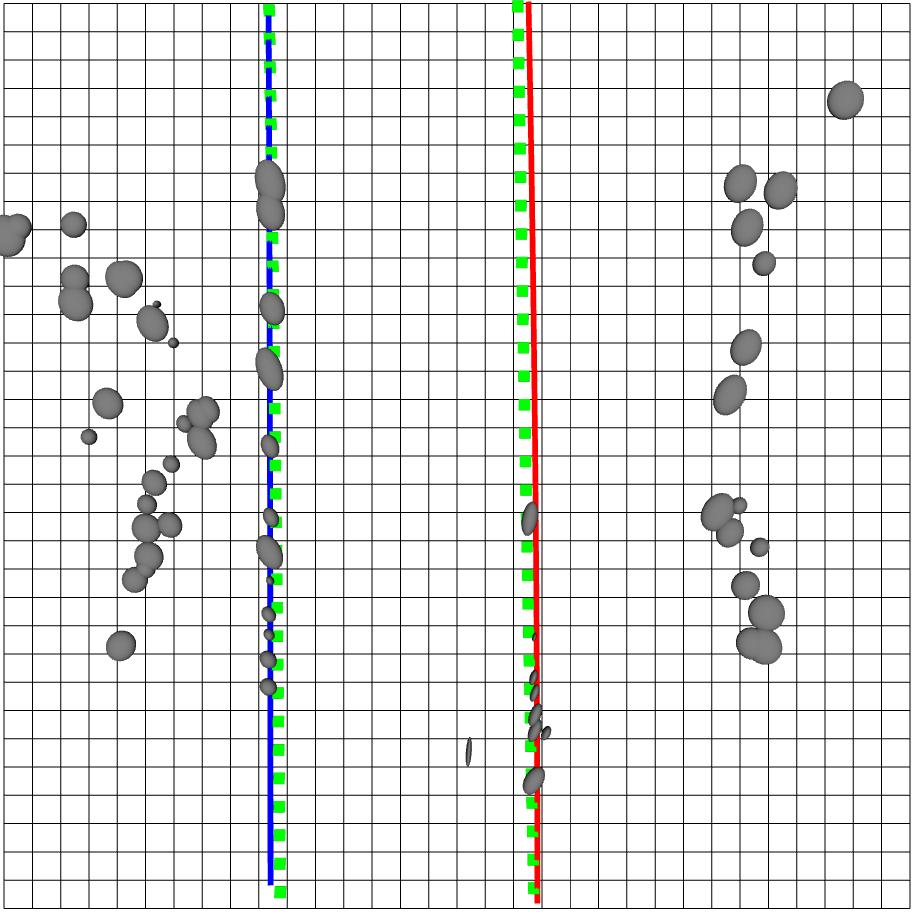}}
\subfigure[]{\includegraphics[width=0.38\linewidth, viewport=116 244 478 544, clip=true]{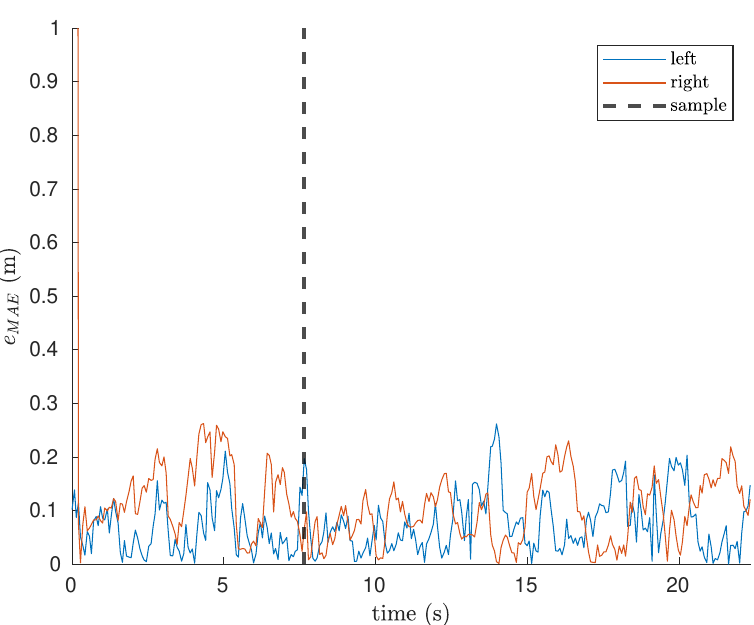}}
\caption{Results from Trajectory 1. (a) Sample camera view and (b) sample top-down orthographic view of scene where grey ellipses are the radar target detections, green squares are ground truth boundary points, and the blue and red curves are the estimated left and right boundaries respectively. Grid has 1 meter spacing. (c) The MAE of the estimated road boundary models vs time, where the dashed line represent the time of the sample in (a) and (b).}
\label{fig:s1}
\end{figure}

\begin{figure}[t!]
\centering
\subfigure[]{\includegraphics[width=0.29\linewidth, viewport=5 5 725 725, clip=true]{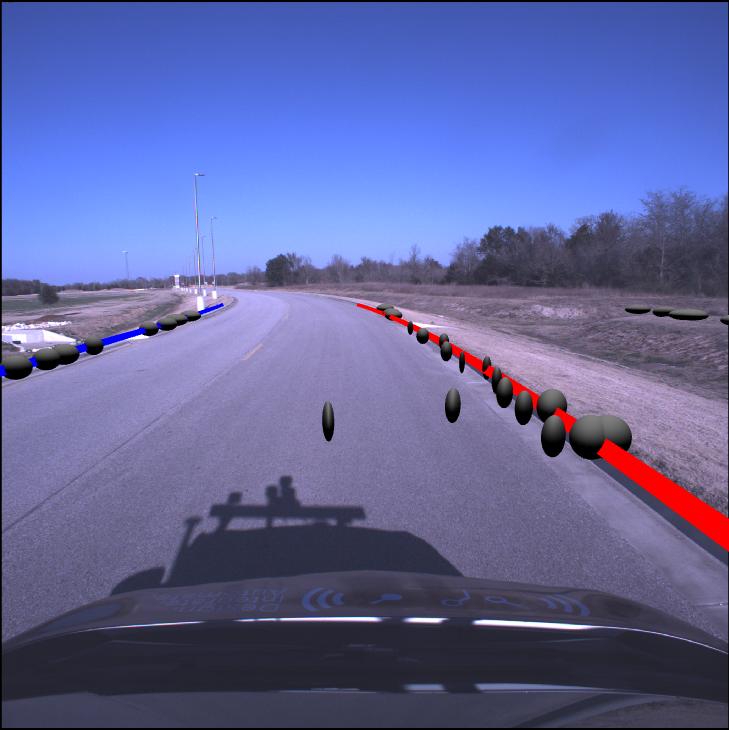}}
\subfigure[]{\includegraphics[width=0.29\linewidth, viewport=0 0 913 912, clip=true]{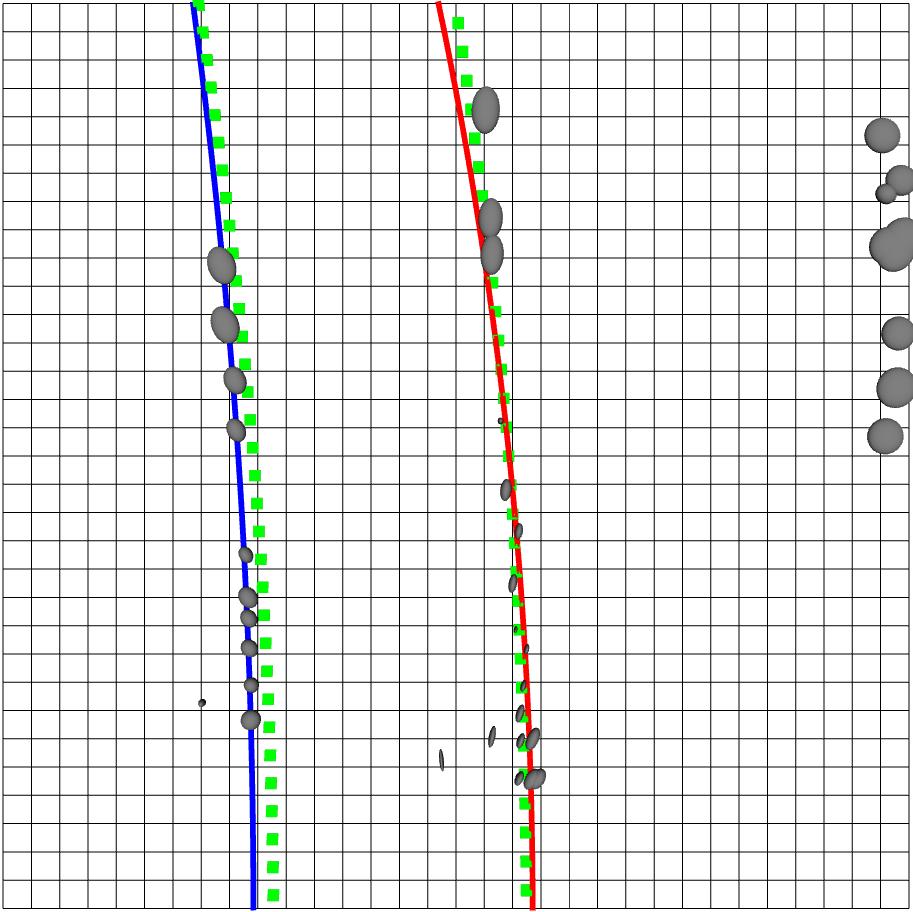}}
\subfigure[]{\includegraphics[width=0.38\linewidth, viewport=116 244 478 544, clip=true]{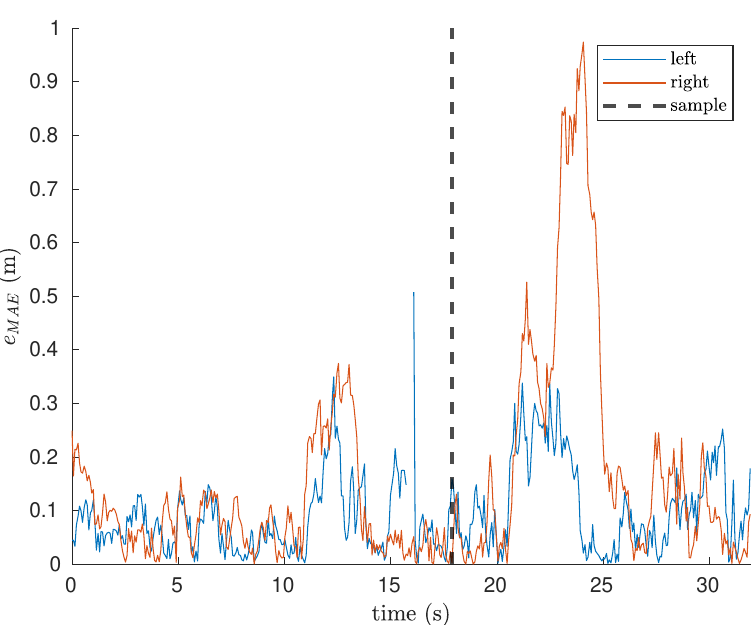}}
\caption{Results from Trajectory 2. (a) Sample camera view. (b) Sample top-down orthographic view. (c) MAE vs time over the trajectory.}
\label{fig:s2}
\end{figure}

\begin{figure}[t!]
\centering
\subfigure[]{\includegraphics[width=0.29\linewidth, viewport=5 5 725 725, clip=true]{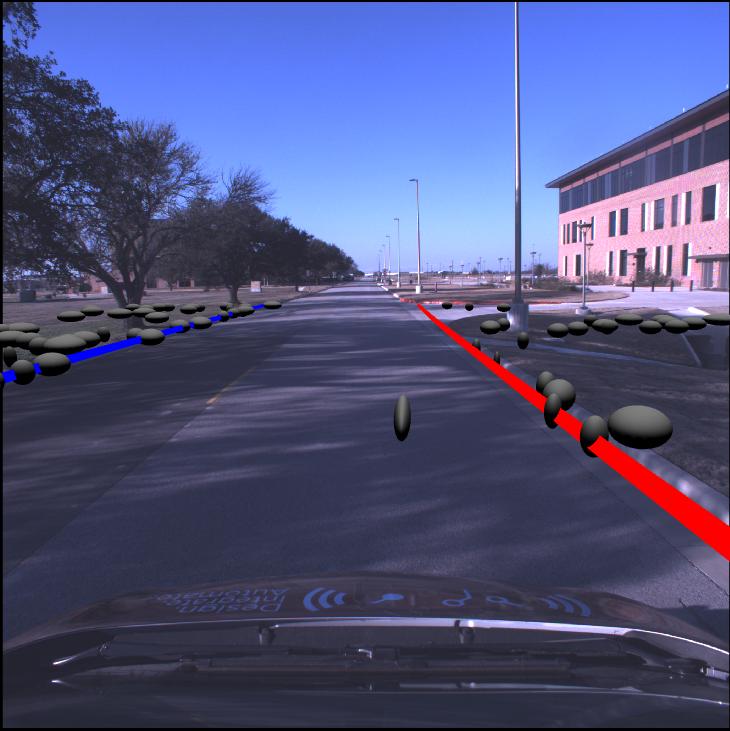}}
\subfigure[]{\includegraphics[width=0.29\linewidth, viewport=0 0 913 912, clip=true]{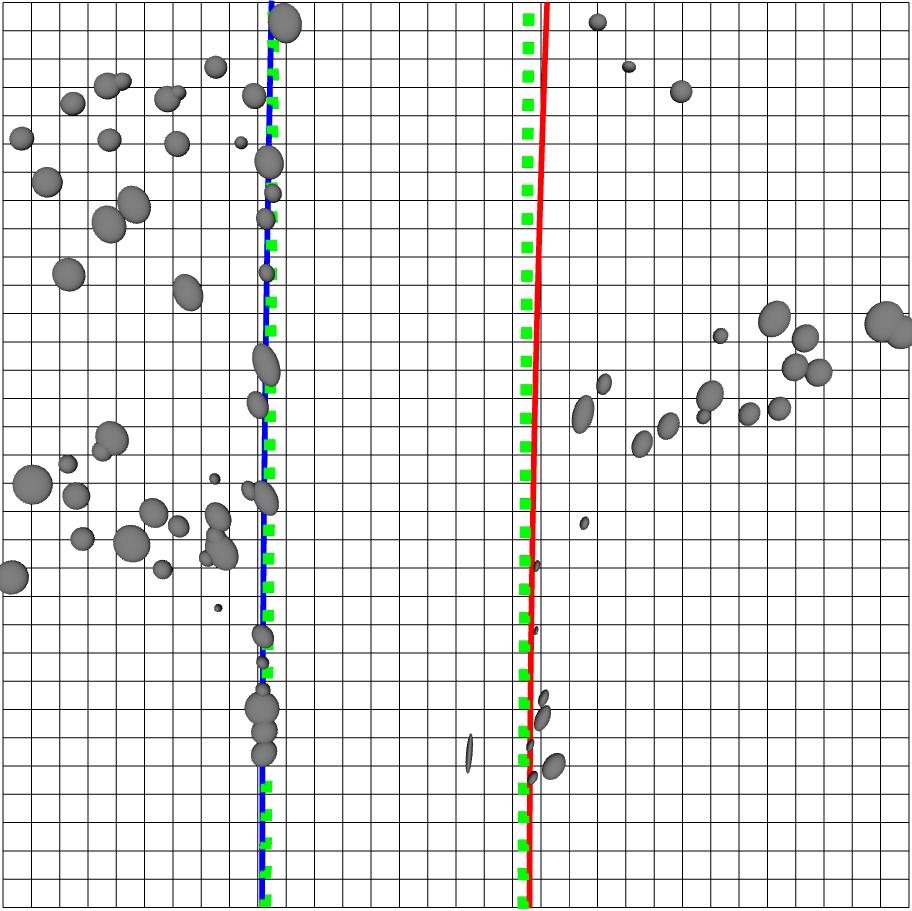}}
\subfigure[]{\includegraphics[width=0.38\linewidth, viewport=116 244 478 544, clip=true]{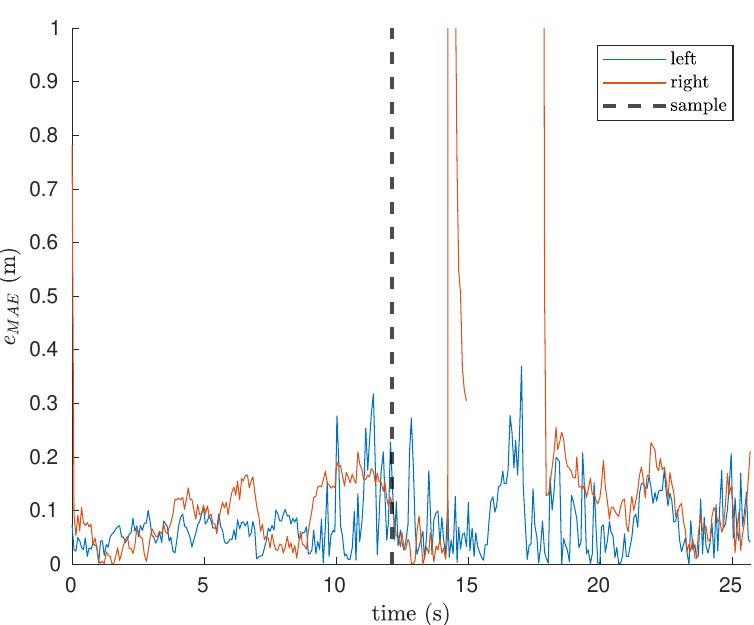}}
\caption{Results from Trajectory 3. (a) Sample camera view. (b) Sample top-down orthographic view. (c) MAE vs time over the trajectory.}
\label{fig:s3}
\end{figure}

\begin{figure*}[t!]
\centering
\includegraphics[width=0.15\linewidth, viewport=5 5 725 725, clip=true]{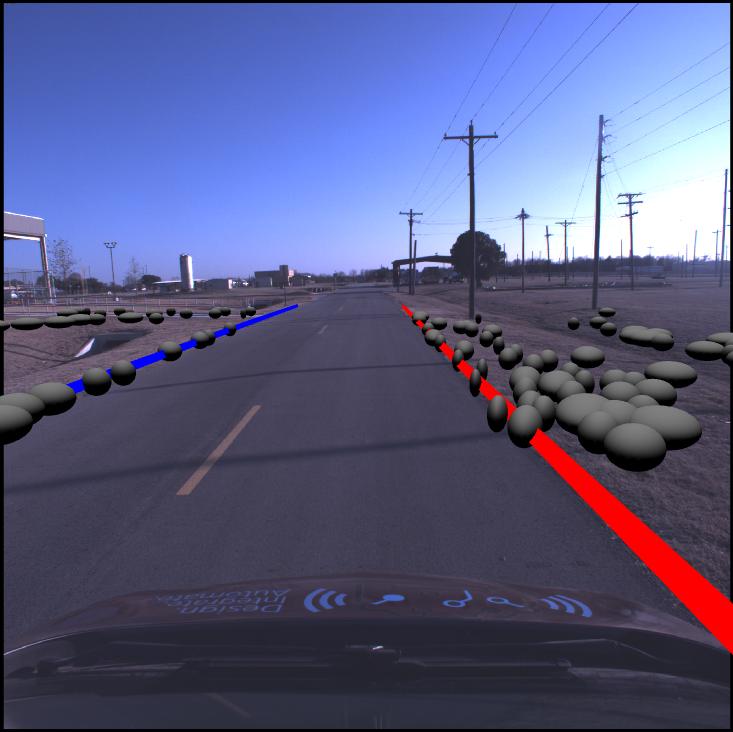}
\includegraphics[width=0.15\linewidth, viewport=0 0 913 912, clip=true]{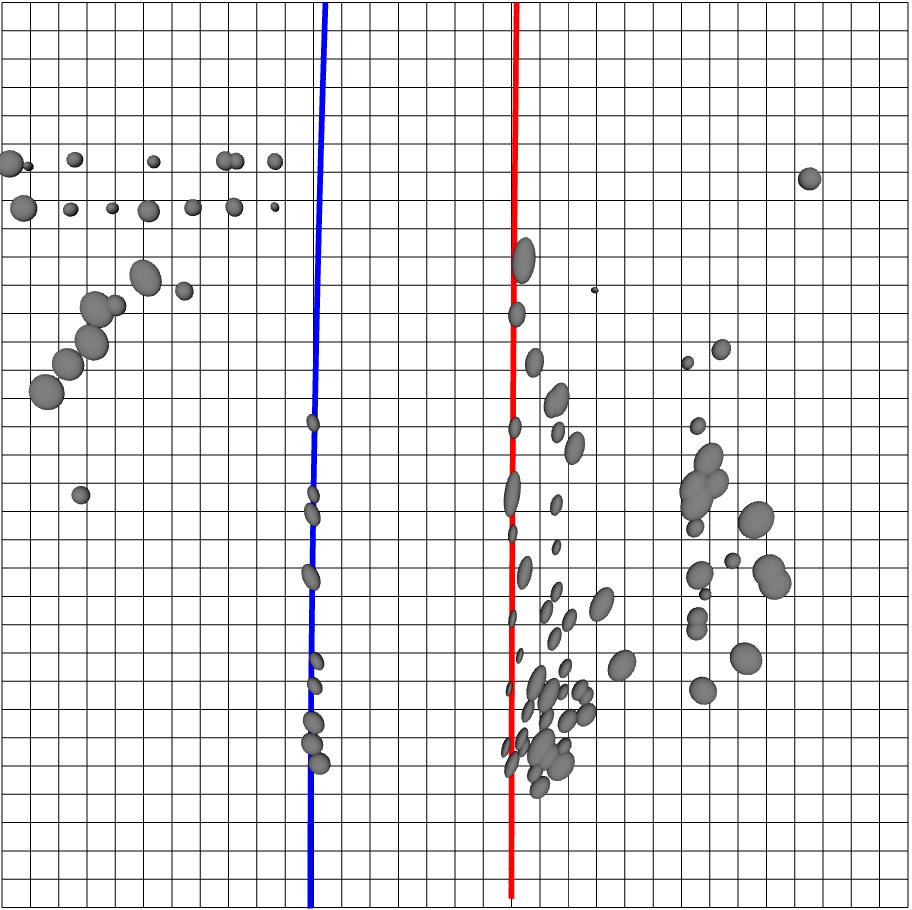}
\includegraphics[width=0.15\linewidth, viewport=5 5 725 725, clip=true]{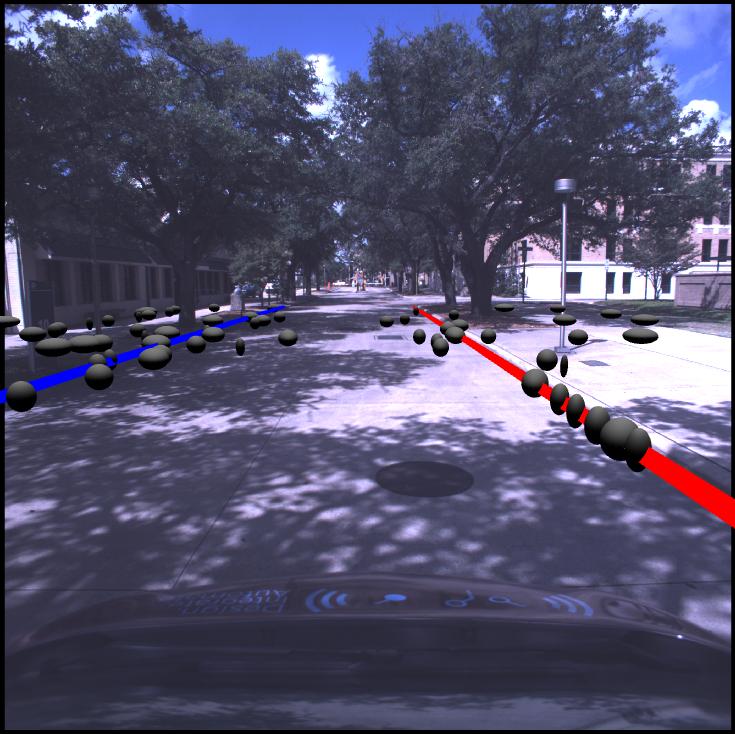}
\includegraphics[width=0.15\linewidth, viewport=0 0 913 912, clip=true]{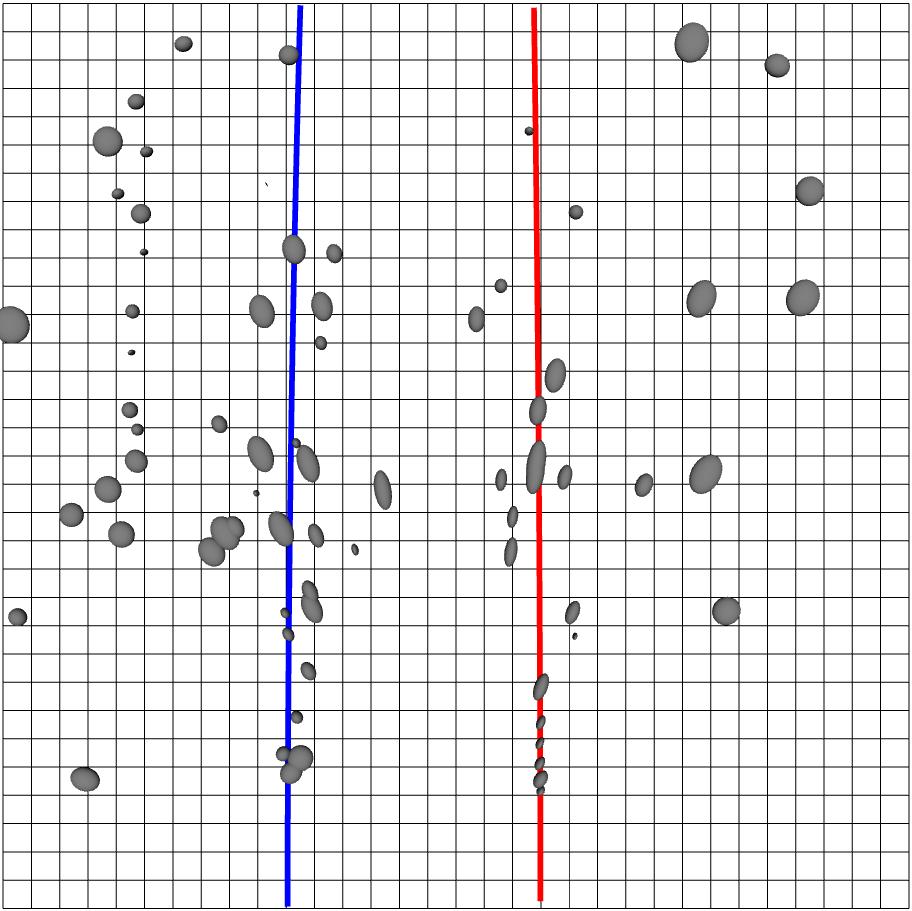}
\includegraphics[width=0.15\linewidth, viewport=5 5 725 725, clip=true]{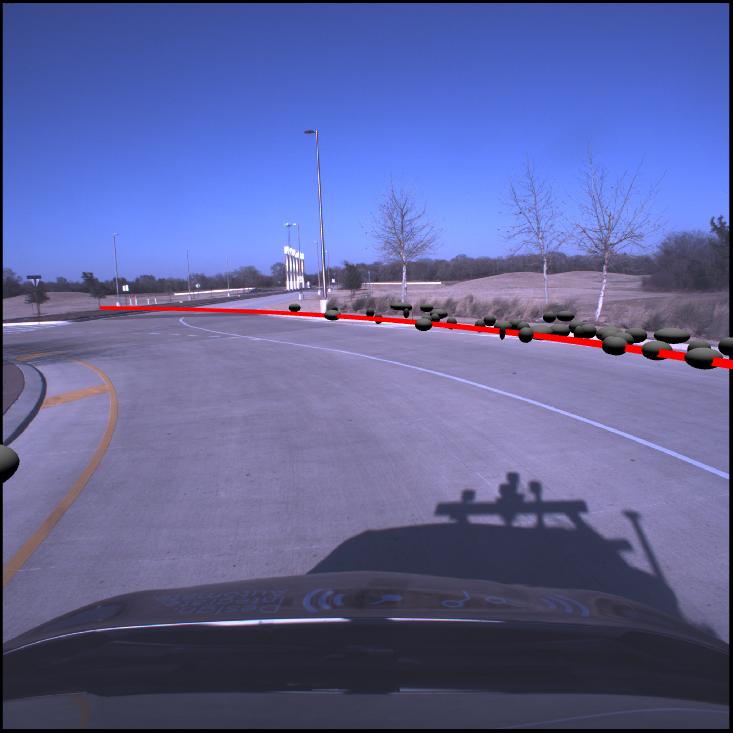}
\includegraphics[width=0.15\linewidth, viewport=0 0 913 912, clip=true]{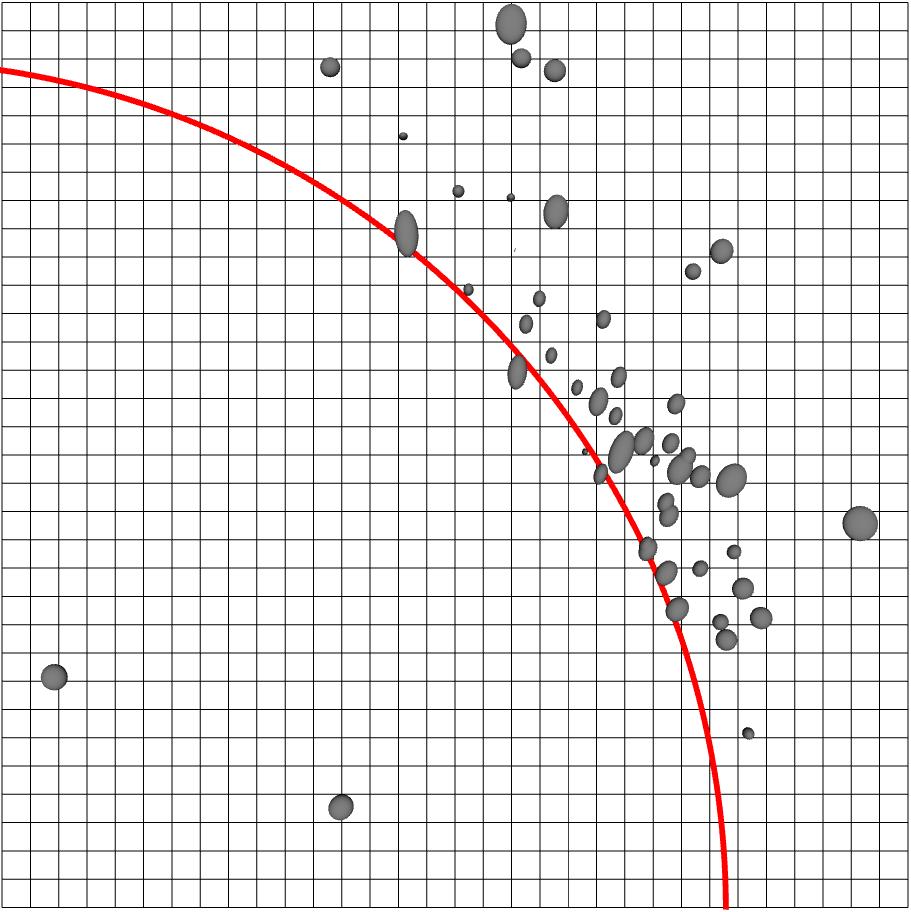}
\caption{A sample of successful road boundary estimations under varying scenarios.}
\label{fig:qual_succ_examples}
\end{figure*}
The proposed algorithm is developed as a Robotic Operating System (ROS) module in C++. We test the algorithm on data collected from a Continental ARS430 automotive Doppler radar, and the algorithm is validated against ground truth measurements of the boundary location represented as discrete points along the boundary with approximately 1 meter spacing. The vehicle drives each trajectory at roughly the posted speed limit of the given roadway.

In order to quantify the results of our algorithm, at each time step we compute the Mean Absolute Error (MAE) of the estimated left and right road boundaries relative to their associated ground truth points. Let us consider a given estimated road boundary $\boldsymbol{\beta}$ and an associated set of ground truth points
\(
\mathbf{S}=\{\mathbf{s}_1,\ldots,\mathbf{s}_N\}
\).
We find that the ground truth points are somewhat conservative, being approximately 20 cm closer to the interior of the roadway than the radar measured points. Therefore, in order to account for this and any GPS offset, we first calculate the mean error over the entire trajectory, which we consider to be the bias due to the ground truth and GPS offsets. The mean error of the estimated boundary is 
\(
e=\frac{1}{N}\sum_{i=1}^N d(\mathbf{s}_i,\boldsymbol{\beta})
\)
where $d(\mathbf{s}_i,\boldsymbol{\beta})$ is the signed geometric distance of the ground truth point $\mathbf{s}_i$ from the boundary $\boldsymbol{\beta}.$ We then compute the mean error $\overline{e}$ and error standard deviation $\sigma_{e}$ over the entire trajectory, excluding any individual estimates where $|e - \overline{e}|>3\sigma_{e},$ which we consider to be failure cases. The mean absolute error of each individual estimate is then $e_{\text{MAE}}=\frac{1}{N}\sum_{i=1}^N |d(\mathbf{s}_i,\boldsymbol{\beta}) - \overline{e}|.$ Additionally, we calculate the failure rate for the estimation of each the left and right boundaries to be the ratio of the number of timesteps where $|e - \overline{e}|>3\sigma_{e}$ or no estimate is given to the total number of timesteps.

We demonstrate three different trajectories where we have ground truth data, for which summarized results are presented in Table \ref{tbl:results_summary}. Trajectory 1, demonstrated in Fig. \ref{fig:s1}, is along a straight, curbed roadway with minimal clutter near the boundary. Similarly, Trajectory 2, demonstrated in Fig. \ref{fig:s2}, is along a curbed roadway with minimal clutter near the boundary, but transitions between straight and curved segments. Trajectory 3, demonstrated in Fig. \ref{fig:s3}, presents a more challenging case along a straight, curbed roadway, featuring clutter on the left roadside, obfuscating the boundary, as well as a driveway intersecting the right boundary.

\begin{table}[!t]
\renewcommand{\arraystretch}{1.3}
\caption{Summarized Results}
\label{tbl:results_summary}
\centering
\begin{tabular}{c|c|c|c|c}
\hline
\bfseries Trajectory & \bfseries Boundary & $\mathbf{\overline{e}_{\text{MAE}}}$ (cm) & $\boldsymbol{\sigma}_{\mathbf{e_\text{MAE}}}$ (cm) & \bfseries \% Failure\\
\hline\hline
1 & left & 7.44 & 8.70 & 0 \\
(300\,m) & right & 10.7 & 6.68 & 0.96 \\
\hline
2 & left & 9.36 & 11.9 & 1.13 \\
(500\,m)  & right & 11.0 & 10.2 & 7.98 \\
\hline
3 & left & 7.50 & 9.17 & 0.56 \\
(220\,m)  & right & 9.98 & 11.7 & 14.0 \\
\hline
\end{tabular}
\end{table}

We consider Trajectory 1 to be the most ideal case for our algorithm, and, consequently, it produces the most stable estimation with no failure cases outside the first few timesteps as the estimation stabilizes. Trajectory 2 results in the lowest estimation accuracy, largely due to the road curvature transitions at $t\approx11\,$s and $t\approx21\,$s which temporarily violate Assumption 3. However, even in this case, the estimation is not considered to fail. For Trajectory 3, the mean MAE is roughly consistent with the ideal case, but with increased variance. Most generally, the estimation accuracy is typically better for the left boundary than the right boundary. This is due to the smaller angle of incidence to the left boundary, providing a greater number of detections. We demonstrate that the algorithm is capable of successful estimation under various challenging conditions, including when there is no physical boundary in Figs. \ref{fig:title_fig} and \ref{fig:qual_succ_examples}.

\begin{figure}[t!]
\centering
\subfigure[]
{
    \includegraphics[width=0.22\linewidth, viewport=5 5 725 725, clip=true]{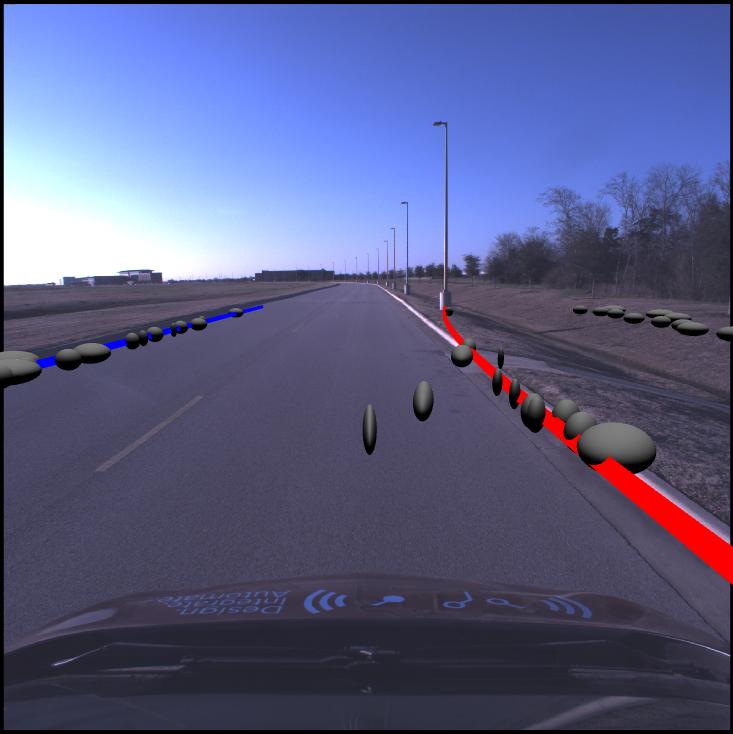}
    \includegraphics[width=0.22\linewidth, viewport=0 0 913 912, clip=true]{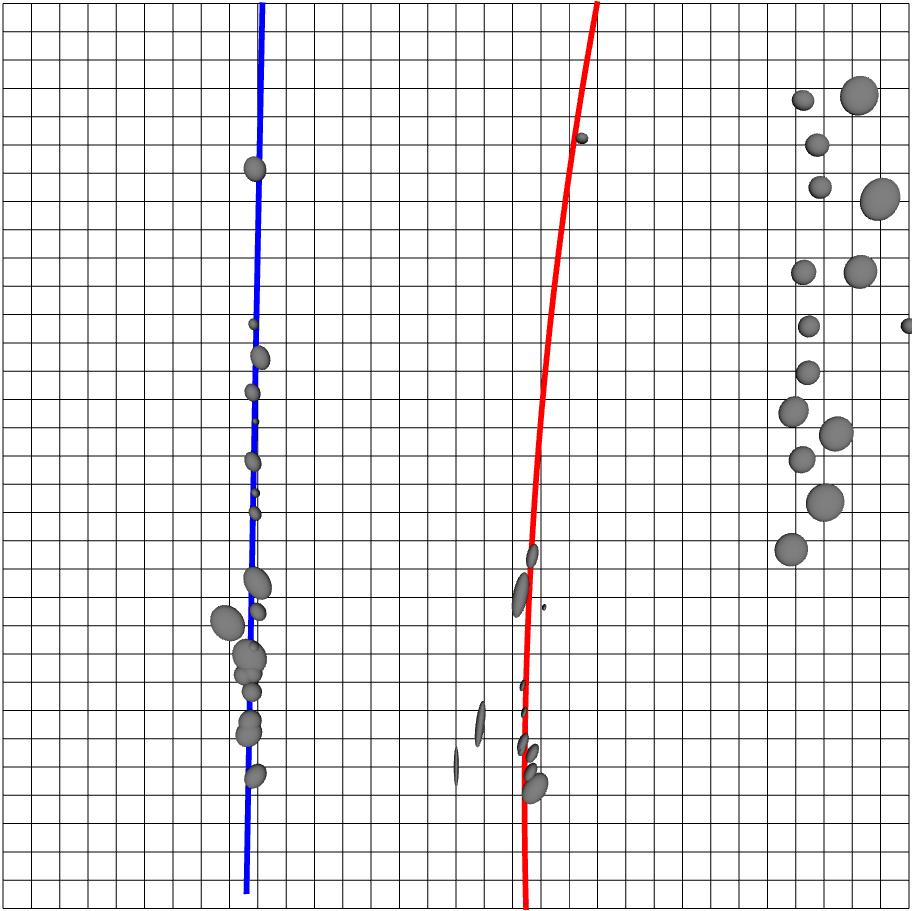}
}
\subfigure[]
{
    \includegraphics[width=0.22\linewidth, viewport=5 5 725 725, clip=true]{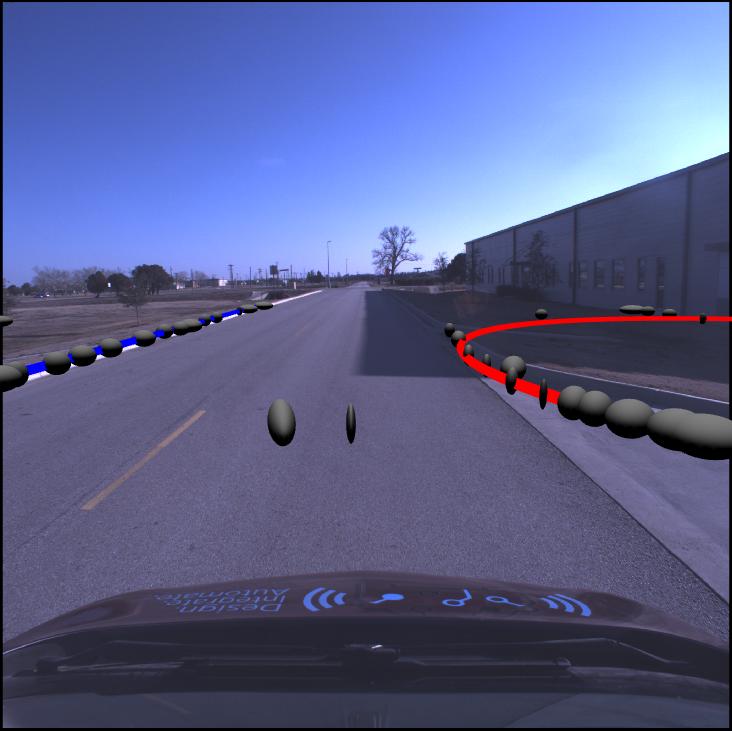}
    \includegraphics[width=0.22\linewidth, viewport=0 0 913 912, clip=true]{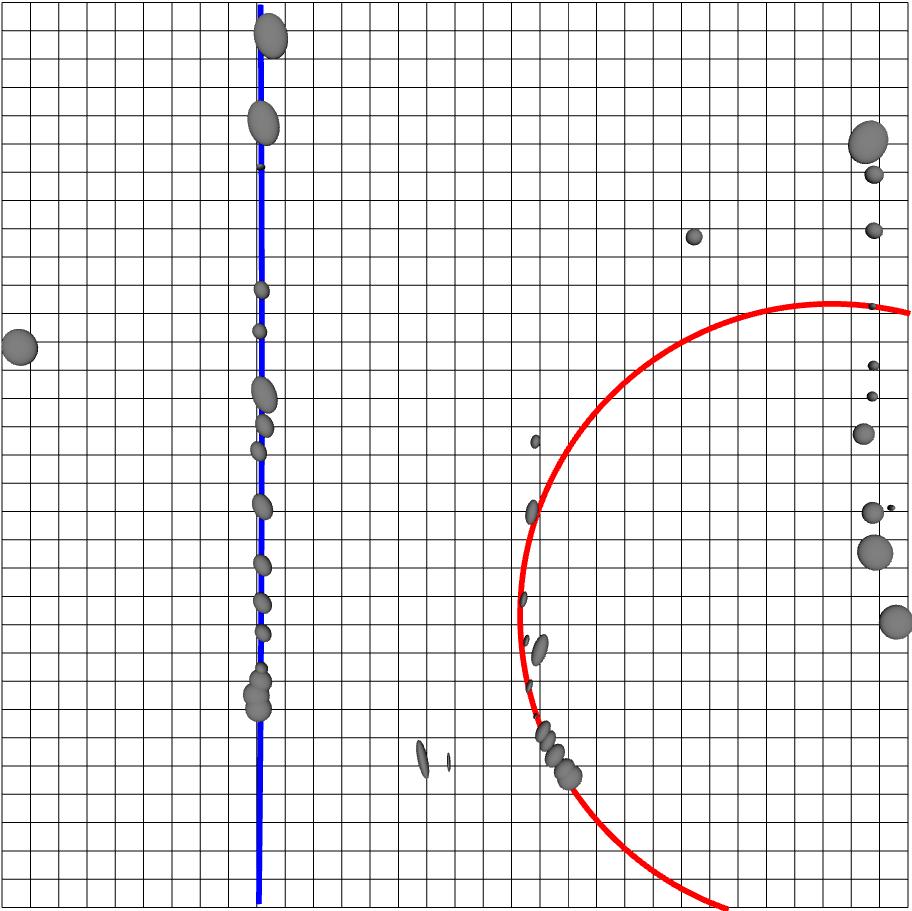}
}
\caption{A sample of common failure modes. (a) High leverage points from clutter can disrupt the boundary estimation. (b) Deviation of boundary at intersection or driveway.}
\label{fig:qual_fail_examples}
\end{figure}
Finally, we identify the common failure modes of the algorithm. The most common, demonstrated in Fig. \ref{fig:qual_fail_examples}(a), occurs when non-boundary detections with high leverage cause the estimation to deviate at long range. This failure mode occurs in Trajectory 2 at $t\approx22\,$s. The other common failure mode, demonstrated in Fig. \ref{fig:qual_fail_examples}(b), occurs when the primary boundary deviates to join with the boundary of an intersecting road or driveway. This failure mode occurs in Trajectory 3 at $t\approx14\,$s. Lastly, the boundary can be occluded by passing vehicles, which occurs once in Trajectory 2 at $t\approx15\,$s when a vehicle passes to the left.

\vspace{-0.7pt}
\section{Conclusion}
We presented a novel road boundary detection method that is solely based on a sparse radar point cloud. The method was built on a homogeneous boundary model and we derived a probability distribution of road boundary models based on the radar point cloud using variational inference. 
In order to generate initial candidate models, we developed a custom RANSAC variant to propose unseen model instances as candidate road boundaries. By alternating variational inference and RANSAC proposals until convergence we generated the best estimate of all candidate models. We selected the candidate model with the minimum lateral distance to the radar on each side as the estimates of the left and right road boundaries.
The algorithm has been implemented as a ROS module and tested under real radar data. The results are satisfactory.

In the future, we will investigate global representations for road boundary mapping and develop a sensor-fusion approach to further improve robustness.

\section*{Acknowledgement}
The authors thank Di Wang, Shuangyu Xie, and Fengzhi Guo for their input and feedback.

\bibliographystyle{ieeetr}
\bibliography{kingery-refs}

\newpage

\onecolumn

\appendices
\section{Joint Probability}
\label{apx:joint_prob}

\begin{figure*}[!h]
\centering
\begin{equation*}
\begin{aligned}
    p(\mathbf{Z},\mathbf{c},\boldsymbol{\pi},\mathbf{B}) &= p(\mathbf{Z}\mid\mathbf{c},\boldsymbol{\pi},\mathbf{B})p(\mathbf{c},\boldsymbol{\pi},\mathbf{B}) \\
    &= p(\mathbf{Z}\mid\mathbf{c},\mathbf{B})p(\mathbf{c}\mid\boldsymbol{\pi},\mathbf{B})p(\boldsymbol{\pi},\mathbf{B}) \\
    &= p(\mathbf{Z}\mid\mathbf{c},\mathbf{B})p(\mathbf{c}\mid\boldsymbol{\pi})p(\boldsymbol{\pi})p(\mathbf{B}) \\
\end{aligned}
\end{equation*}

where
\begin{equation*}
\begin{aligned}
    p(\mathbf{Z}\mid\mathbf{c},\mathbf{B}) &= &&\prod_{i=1}^N (\mathcal{U}(\mathbf{z}_i;[0,r_{\max}]\times[\theta_{\min},\theta_{\max}])^{[c_i = 0]} \\
    &&& \prod_{k=1}^K \mathcal{N}(\boldsymbol{\beta}_k^\mathsf{T}\boldsymbol{\phi}(\mathbf{z}_i); \mathbf{0}, \sigma_{ik})^{[c_i=k]})\\
    p(\mathbf{c}\mid\boldsymbol{\pi}) &= &&\prod_{i=1}^N \prod_{k=0}^K \pi_k^{[c_i=k]} \\
    p(\boldsymbol{\pi}) &= &&\frac{\Gamma(\sum_{k=0}^K\alpha_k)}{\prod_{k=0}^K\Gamma(\alpha_k)} \prod_{k=0}^K \pi_k^{\alpha_k - 1} \\
    p(\mathbf{B}) &= &&\prod_{k=1}^K \mathcal{B}(\boldsymbol{\beta}_k;\mathbf{C}_k^{-1})
\end{aligned}
\end{equation*}
\end{figure*}

\newpage
\section{Derivation of Expectation Step}
\label{apx:estep_deriv}
\begin{figure*}[!h]
\centering
\begin{equation*}
\begin{aligned}
    \log q^*(\mathbf{c}) &=&& E_{\boldsymbol{\pi},\mathbf{B}}[\log p(\mathbf{Z}\mid\mathbf{c},\mathbf{B}) + \log p(\mathbf{c}\mid\boldsymbol{\pi})] + \text{constant}\\
    &=&& E_{\mathbf{B}}[\log p(\mathbf{Z}\mid\mathbf{c},\mathbf{B})] + E_{\boldsymbol{\pi}}[\log p(\mathbf{c}\mid\boldsymbol{\pi})] + \text{constant}\\
    &=&& E_{\mathbf{B}}\left[\log \prod_{i=1}^N \left(\mathcal{U}(\mathbf{z}_i;[0,r_{\max}]\times[\theta_{\min},\theta_{\max}])^{[c_i = 0]}\prod_{k=1}^K \mathcal{N}(\boldsymbol{\beta}_k^\mathsf{T}\boldsymbol{\phi}(\mathbf{z}_i); \mathbf{0}, \sigma_{ik})^{[c_i=k]}\right)\right] \\
      &&& + E_{\boldsymbol{\pi}}\left[\log \prod_{i=1}^N \prod_{k=0}^K \pi_k^{[c_i=k]}\right] + \text{constant} \\
    &=&& \sum_{i=1}^N \left([c_i=0]\log\mathcal{U}(\mathbf{z}_i;[0,r_{\max}]\times[\theta_{\min},\theta_{\max}]) + \sum_{k=1}^K [c_i=k]E_{\mathbf{B}}\left[\log\mathcal{N}(\boldsymbol{\beta}_k^\mathsf{T}\boldsymbol{\phi}(\mathbf{z}_i); \mathbf{0}, \sigma_{ik})\right]\right) \\ 
      &&& + \sum_{i=1}^N \sum_{k=0}^K [c_i=k]E_{\boldsymbol{\pi}}\left[\log \pi_k\right] + \text{constant} \\
    &=&& \sum_{i=1}^N [c_i=0]\log\mathcal{U}(\mathbf{z}_i;[0,r_{\max}]\times[\theta_{\min},\theta_{\max}]) \\
      &&& + \sum_{i=1}^N\sum_{k=1}^K [c_i=k]E_{\mathbf{B}}\left[\log\mathcal{N}(\boldsymbol{\beta}_k^\mathsf{T}\boldsymbol{\phi}(\mathbf{z}_i); \mathbf{0}, \sigma_{ik})\right] \\
      &&& + \sum_{i=1}^N \sum_{k=0}^K [c_i=k]E_{\boldsymbol{\pi}}\left[\log \pi_k\right] + \text{constant} \\
    &=&& -\sum_{i=1}^N [c_i=0](\log(r_{\max})+\log(\theta_{\max}-\theta_{\min})) \\
      &&& - \sum_{i=1}^N\sum_{k=1}^K [c_i=k]\left(\frac{1}{2}\log(2\pi)+\log(\sigma_{ik})+\frac{1}{2}E_{\mathbf{B}}\left[\frac{\boldsymbol{\beta}_k^\mathsf{T}\boldsymbol{\phi}(\mathbf{z}_i)\boldsymbol{\phi}(\mathbf{z}_i)^\mathsf{T}\boldsymbol{\beta}_k}{\sigma_{ik}^2}\right]\right) \\
      &&& + \sum_{i=1}^N \sum_{k=0}^K [c_i=k]E_{\boldsymbol{\pi}}\left[\log \pi_k\right] + \text{constant}
\end{aligned}
\end{equation*}
Let
\begin{equation*}
\begin{aligned}
\log\rho_{ik} &= E_{\boldsymbol{\pi}}\left[\log \pi_k\right] -
    \begin{cases}
        \log(r_{\max})+\log(\theta_{\max}-\theta_{\min}) & k=0 \\
        \frac{1}{2}\log(2\pi)+\log(\sigma_{ik})+\frac{1}{2}E_{\mathbf{B}}\left[\frac{\boldsymbol{\beta}_k^\mathsf{T}\boldsymbol{\phi}(\mathbf{z}_i)\boldsymbol{\phi}(\mathbf{z}_i)^\mathsf{T}\boldsymbol{\beta}_k}{\sigma_{ik}^2}\right] & k=1,\ldots,K
    \end{cases} \\
\rho_{ik} &= \pi_k 
    \begin{cases}
        \mathcal{U}(\mathbf{z}_i;[0,r_{\max}]\times[\theta_{\min},\theta_{\max}]) & k = 0 \\
        \mathcal{N}(\boldsymbol{\beta}_k^\mathsf{T}\boldsymbol{\phi}(\mathbf{z}_i); \mathbf{0}, \sigma_{ik}) & k = 1,\ldots,K
    \end{cases}
\end{aligned}
\end{equation*}
Then
\begin{equation*}
\begin{aligned}
    \log q^*(\mathbf{c}) &= \sum_{i=1}^N \sum_{k=0}^K [c_i=k] \log\rho_{ik} + \text{constant} \\
    q^*(\mathbf{c}) &\propto \prod_{i=1}^N \prod_{k=0}^K \rho_{ik}^{[c_i=k]} \\
    \gamma_{ik} &= \frac{\rho_{ik}}{\sum_{j=0}^K \rho_{ij}} \\
    q^*(\mathbf{c}) &= \prod_{i=1}^N \prod_{k=0}^K \gamma_{ik}^{[c_i=k]} 
\end{aligned}
\end{equation*}
Therefore, $c_i\sim\text{Categorical}(K+1,\boldsymbol{\gamma}_i)$ and $E_\mathbf{c}[\,[c_i=k]\,]=\gamma_{ik}$.
\end{figure*}

\section{Derivation of Maximization Step}
\label{apx:mstep_deriv}
\begin{figure*}[!h]
\centering
\begin{equation*}
\begin{aligned}
    \log q^*(\boldsymbol{\pi}) &= E_{\mathbf{c}}[\log p(\mathbf{c}\mid\boldsymbol{\pi}) + \log p(\boldsymbol{\pi})] + \text{constant} \\
    &=  E_{\mathbf{c}}[\log p(\mathbf{c}\mid\boldsymbol{\pi})] + \log p(\boldsymbol{\pi}) + \text{constant} \\
    &=  E_{\mathbf{c}}[\log \prod_{i=1}^N \prod_{k=0}^K \pi_k^{[c_i=k]}] + \log \frac{\Gamma(\sum_{k=0}^K\alpha_k)}{\prod_{k=0}^K\Gamma(\alpha_k)} \prod_{k=0}^K \pi_k^{\alpha_k - 1} + \text{constant} \\
    &= \sum_{i=1}^N \sum_{k=0}^K E_{\mathbf{c}}[\,[c_i=k]\,]\log\pi_k + \log \Gamma\left(\sum_{k=0}^K\alpha_k\right) - \sum_{k=0}^K\log\Gamma(\alpha_k)  + \sum_{k=0}^K (\alpha_k - 1)\log\pi_k + \text{constant} \\
    &= \sum_{i=1}^N \sum_{k=0}^K \gamma_{ik}\log\pi_k +\sum_{k=0}^K (\alpha_k - 1)\log\pi_k + \text{constant} \\
    q^*(\boldsymbol{\pi}) &\propto \prod_{k=0}^K \pi_k^{\alpha_k - 1 + \sum_{i=1}^N \gamma_{ik}}
\end{aligned}
\end{equation*}
Therefore, $\boldsymbol{\pi}\sim\text{Dirichlet}(\boldsymbol{\alpha}'\mid\alpha'_k = \alpha_k + \sum_{i=1}^N\gamma_{ik})$ and $E_{\boldsymbol{\pi}}[\pi_k]=\frac{\alpha'_k}{\sum_{k=0}^K \alpha'_k}$.

\begin{equation*}
\begin{aligned}
    \log q^*(\mathbf{B}) &=&& E_{\mathbf{c}}[\log p(\mathbf{Z}\mid\mathbf{c},\mathbf{B}) + \log p(\mathbf{B})] + \text{constant} \\
    &=&& E_{\mathbf{c}}[\log p(\mathbf{Z}\mid\mathbf{c},\mathbf{B})] + \log p(\mathbf{B}) + \text{constant} \\
    &=&& E_{\mathbf{c}}[\log \prod_{i=1}^N (\mathcal{U}(\mathbf{z}_i;[0,r_{\max}]\times[\theta_{\min},\theta_{\max}])^{[c_i = 0]} 
    \prod_{k=1}^K \mathcal{N}(\boldsymbol{\beta}_k^\mathsf{T}\boldsymbol{\phi}(\mathbf{z}_i); \mathbf{0}, \sigma_{ik})^{[c_i=k]})] \\
      &&& + \log \prod_{k=1}^K \mathcal{B}(\boldsymbol{\beta}_k;\mathbf{C}_k^{-1}) + \text{constant} \\
    &=&& \sum_{i=1}^N E_{\mathbf{c}}[\,[c_i = 0]\,]\log\mathcal{U}(\mathbf{z}_i;[0,r_{\max}]\times[\theta_{\min},\theta_{\max}]) \\
      &&& + \sum_{i=1}^N \sum_{k=1}^K E_{\mathbf{c}}[\,[c_i=k]\,]\log\mathcal{N}(\boldsymbol{\beta}_k^\mathsf{T}\boldsymbol{\phi}(\mathbf{z}_i); \mathbf{0}, \sigma_{ik}) \\
      &&& + \sum_{k=1}^K \log \mathcal{B}(\boldsymbol{\beta}_k;\mathbf{C}_k^{-1}) + \text{constant} \\
    &=&& \sum_{i=1}^N \sum_{k=1}^K \gamma_{ik}\log\mathcal{N}(\boldsymbol{\beta}_k^\mathsf{T}\boldsymbol{\phi}(\mathbf{z}_i); \mathbf{0}, \sigma_{ik}) + \sum_{k=1}^K \log \mathcal{B}(\boldsymbol{\beta}_k;\mathbf{C}_k^{-1}) + \text{constant} \\
    &=&& -\sum_{i=1}^N \sum_{k=1}^K \gamma_{ik}\left(\frac{1}{2}\frac{\boldsymbol{\beta}_k^\mathsf{T}\boldsymbol{\phi}(\mathbf{z}_i)\boldsymbol{\phi}(\mathbf{z}_i)^\mathsf{T}\boldsymbol{\beta}_k}{\sigma_{ik}^2}\right) -\sum_{k=1}^K \frac{1}{2}\boldsymbol{\beta}_k^\mathsf{T}\mathbf{C}_k^{-1}\boldsymbol{\beta}_k + \text{constant} \\
    &=&& -\frac{1}{2}\sum_{i=1}^N \sum_{k=1}^K \gamma_{ik}\frac{\boldsymbol{\beta}_k^\mathsf{T}\boldsymbol{\phi}(\mathbf{z}_i)\boldsymbol{\phi}(\mathbf{z}_i)^\mathsf{T}\boldsymbol{\beta}_k}{\sigma_{ik}^2}
      - \frac{1}{2}\sum_{k=1}^K \boldsymbol{\beta}_k^\mathsf{T}\mathbf{C}_k^{-1}\boldsymbol{\beta}_k + \text{constant} \\
    &=&& -\frac{1}{2}\sum_{k=1}^K\boldsymbol{\beta}_k^\mathsf{T}\left(\mathbf{C}_k^{-1} + \sum_{i=1}^N\gamma_{ik}\frac{\boldsymbol{\phi}(\mathbf{z}_i)\boldsymbol{\phi}(\mathbf{z}_i)^\mathsf{T}}{\sigma_{ik}^2}\right)\boldsymbol{\beta}_k \\
    q^*(\mathbf{B}) &\propto&& \prod_{k=1}^K \exp\left(-\frac{1}{2}\boldsymbol{\beta}_k^\mathsf{T}\left(\mathbf{C}_k^{-1} + \sum_{i=1}^N\gamma_{ik}\frac{\boldsymbol{\phi}(\mathbf{z}_i)\boldsymbol{\phi}(\mathbf{z}_i)^\mathsf{T}}{\sigma_{ik}^2}\right)\boldsymbol{\beta}_k\right)
\end{aligned}
\end{equation*}
Therefore $\boldsymbol{\beta}_k\sim\text{Bingham}\left(\mathbf{C}_k^{-1} + \sum_{i=1}^N\gamma_{ik}\frac{\boldsymbol{\phi}(\mathbf{z}_i)\boldsymbol{\phi}(\mathbf{z}_i)^\mathsf{T}}{\sigma_{ik}^2}\right)$. Note that $\sigma_{ik}^2=\boldsymbol{\beta}_k^\mathsf{T}\boldsymbol{\Phi}(\mathbf{z}_i)\boldsymbol{\Sigma}_{\mathbf{z},i}\boldsymbol{\Phi}(\mathbf{z}_i)^\mathsf{T}\boldsymbol{\beta}_k$. 
\end{figure*}
\end{document}